\documentclass[conference]{IEEEtran}
\IEEEoverridecommandlockouts
\usepackage{cite}
\usepackage{amsmath,amssymb,amsfonts}
\usepackage{algorithmic}
\usepackage{hyperref} 
\usepackage{subcaption} 
\usepackage{multirow}
\usepackage{graphicx}
\usepackage{textcomp}
\usepackage{xcolor}
\def\BibTeX{{\rm B\kern-.05em{\sc i\kern-.025em b}\kern-.08em
    T\kern-.1667em\lower.7ex\hbox{E}\kern-.125emX}}
\begin{document}

\title{EMMI - Empathic Multimodal Motivational Interviews Dataset: Analyses and Annotations\\
 \thanks{ANR Tapas}
}

 \author{\IEEEauthorblockN{Lucie Galland}
\IEEEauthorblockA{\textit{ISIR} \\
 \textit{Sorbonne university}\\
 Paris, France \\
 lucie.galland@isir.upmc.fr}
 \and
 \IEEEauthorblockN{Catherine Pelachaud}
\IEEEauthorblockA{\textit{CNRS, ISIR} \\
 \textit{Sorbonne university}\\
Paris, France \\
 catherine.pelachaud@isir.upmc.fr}
 \and
 \IEEEauthorblockN{Florian Pecune}
 \IEEEauthorblockA{\textit{CNRS - SANPSY, } \\
 \textit{Bordeaux University}\\
 Bordeaux, France \\
 florian.pecune@u-bordeaux.fr}
 }

\maketitle

\begin{abstract}
The study of multimodal interaction in therapy can yield a comprehensive understanding of therapist and patient behavior that can be used to develop a multimodal virtual agent supporting therapy. This investigation aims to uncover how therapists skillfully blend therapy's task goal (employing classical steps of Motivational Interviewing) with the social goal (building a trusting relationship and expressing empathy). Furthermore, we seek to categorize patients into various ``types'' requiring tailored therapeutic approaches. To this intent, we present multimodal annotations of a corpus consisting of simulated motivational interviewing conversations, wherein actors portray the roles of patients and therapists. We introduce EMMI, composed of two publicly available MI corpora, AnnoMI and the Motivational Interviewing Dataset, for which we add multimodal annotations. We analyze these annotations to characterize functional behavior for developing a virtual agent performing motivational interviews emphasizing social and empathic behaviors. Our analysis found three clusters of patients expressing significant differences in behavior and adaptation of the therapist's behavior to those types. This shows the importance of a therapist being able to adapt their behavior depending on the current situation within the dialog and the type of user.
\end{abstract}


\section{Introduction}

The prevalence of mental health issues has witnessed a notable uptick, leading to a substantial disparity between the demand for mental health services and the available resources  \cite{cameron2017towards}. Consequently, patients frequently encounter prolonged wait times before they can commence therapy  \cite{cameron2017towards,denecke2020mental}. To tackle this predicament, a potential remedy emerges in the form of virtual agents designed to replicate Motivational Interviews (MI) for patients awaiting available appointments. These virtual agents hold promise in alleviating the waiting predicament by offering instantaneous support% and interventions
, particularly within therapeutic modalities involving multiple sessions \cite{fiske2019your}. MI is a therapeutic approach underscored by collaboration and the encouragement of behavioral transformation. During Motivational Interviews, therapists employ various strategies to guide patients toward articulating their motivation for change. These strategies can be composed of verbal (reflection, question) and nonverbal elements (smiles, head and body position). A virtual agent using both verbal and nonverbal strategies to conduct motivational interviews could possess the capacity to rekindle the user's motivation whenever necessary.
Nonetheless, creating such an agent necessitates the examination of Human-Human data, which proves challenging to obtain due to the sensitive nature of the subjects discussed. Furthermore, delving into the multimodal conduct of therapists would provide invaluable insights into the agent's development, given that psychological theory underscores the significance of nonverbal behavior during such interviews \cite{gillam2019brief}. Another aspect highlighted by psychological theory is the pivotal role of empathic and social conduct. Patients are more likely to be honest and share their symptoms and concerns with an empathic physician, leading to better therapy outcomes \cite{jani2012role}. Clinical empathy implies perceiving how the patient is experiencing the world and how the patient is currently feeling.

In this paper, we present a series of multimodal annotations and analyses carried out on two existing MI corpora, namely AnnoMI \cite{wu2023creation}, and the Motivational Interviewing Dataset (MID)  \cite{perez2019makes}. These annotations aim to scrutinize the empathic conduct of therapists during MI sessions, along with the behaviors and responses of patients to such empathic interactions. These annotations form the Empathic Multimodal Motivational Interviews (EMMI) dataset.

This paper offers the following contributions. First, we perform multimodal annotations of publicly available MI corpora, creating EMMI. Then, we analyze patients' behaviors, highlighting three different types of patients. Finally, we analyze therapists' multimodal behavior depending on the type of patient they interact with.

\paragraph{Research Questions}

The study of multimodal interaction in therapy can yield a comprehensive understanding of therapist and patient behavior that can be used to develop a multimodal virtual agent supporting therapy. Currently, most datasets in this field primarily focus on the verbal aspect, only considering limited information from nonverbal behavior. Existing studies often concentrate on comparing high- and low-quality interactions, neglecting a deeper analysis of social behaviors. Our research analyzes therapy session videos, considering the patients' and therapists' progression on verbal and nonverbal levels and their alignment. This investigation aims to address such questions and gain insights into the interactions and patterns between the therapist and the patient. Our primary objective is to discern how therapists blend their task goal (employing classical steps of MI) with their social goal (building a trusting relationship and expressing empathy). We also examine how these goals manifest in their verbal and nonverbal behaviors. Furthermore, this study aims to uncover how patients respond to different therapeutic strategies employed by therapists and identify cues indicative of positive effects on patients, potentially leading to categorizing patients into various ``types'' requiring tailored therapeutic approaches. Lastly, we strive to determine the reciprocal influence of behaviors between the two interlocutors during the therapeutic process. 

We address the following research questions:

\textbf{RQ1:} Do the behavior of the therapist and the patient evolve throughout the interaction?

\textbf{RQ2: }Can distinct ``types'' of patients be identified regarding the change in talk patterns, and to what extent do they lead to differences in patients' and therapists' behavior? 

\textbf{RQ3:} How do therapists adapt their verbal and nonverbal behavior to the talk types of patients?

By investigating these research questions, we aim to gain insights into multimodal interaction during therapy and contribute to developing effective and personalized virtual therapeutic interventions.

In the next section, MI is defined, existing MI corpora are presented, and EMMI is presented. Section \ref{sec:annot} describes our annotation process and introduces different types of MI patients. In Section \ref{sec:analysis}, we analyze our annotations to highlight differences in the behaviors of patients and therapists depending on their type, change talk, or whether it is the first or the second half of the conversation. Finally, in Section\ref{sec:discussion}, we interpret these differences and discuss the impact of types on both the patient's and therapists' behavior.
\section{Background and Existing Corpora}
\paragraph{Motivational interviewing}
Motivational Interviewing (MI) is an approach to therapy that emphasizes collaboration and encourages behavioral change.
During MI, therapists use strategies to guide patients toward expressing motivation toward change \cite{miller2012motivational}. One of the main strategies is using reflections, where therapists convey their understanding of what the patient is saying or feeling without judging, interpreting, or advising. Patient behaviors are classically coded into three categories defined by the Motivational Interviewing Skill Code (MISC) \cite{miller2003manual} with \textbf{Change talk (CT)} reflecting actions toward behavior change, \textbf{Sustain talk (ST)}: reflecting actions away from behavior change, \textbf{Follow/Neutral (F/N)}: unrelated to the target behavior.
\paragraph{Existing corpora}

Numerous studies have delved into exploring MI at the verbal level. However, a significant challenge arises from the unavailability of relevant data due to the sensitive nature of the topics discussed within MI conversations. Most Human-Human MI corpora cannot be publicly disseminated or are privately owned. For instance, research by \cite{perez2016building,perez2017predicting} analyzes therapist behavior using phone conversation transcripts of actual therapy sessions, but these datasets cannot be shared. \cite{perez2017understanding} collected a private corpus of audio-recorded therapy session interactions to study therapist empathy. Similarly, \cite{tavabi2020multimodal} focuses on patient behavior classification using transcribed audio data, with only publicly available text transcripts. 

The preceding corpora are based on audio recording, making the study of nonverbal behaviors impossible. Although some video-based corpora have emerged, such as the \href{https://alexanderstreet.com/products/counseling-therapy-video-library}{Counseling and Therapy Video Library}, they are privately owned and often can be acquired at a high cost. Only a few corpora (not publicly available) exist to study facial expressions and body language; for example, \cite{nakano2022detecting} assembles a video corpus for detecting change talk, but it is inaccessible.
On the other hand, there have been recent efforts to create publicly available datasets. \cite{perez2019makes} introduces a corpus of MI videos scraped from the web, though automatic YouTube captioning leads to transcription errors.
Additionally, \cite{wu2023creation} provides a publicly accessible MI video corpus transcribed and annotated by experts. ANNOMI comprises high-quality interactions designed to teach good practices in MI and low-quality interactions designed to show mistakes to avoid during an MI intervention. These studies primarily focus on text analysis, particularly the behavior of therapists of varying qualities. 

However, none of these publicly available resources delve into the video and audio modalities, nor do they explore how patients' behavior varies during a session and how therapist adapts their strategies during the interaction. This paper addresses this gap by presenting a comprehensive development and analysis of a multimodal therapist behavior dataset. The dataset is collected from online open sources, making it accessible to the broader research community.
\paragraph{EMMI}
EMMI comprises 285 videos, representing 21 hours and 22 minutes, complementing two pre-existing datasets: AnnoMI \cite{wu2023creation} and the MID \cite{perez2019makes}. These datasets consist of simulated motivational interview (MI) conversations in which actors portray the roles of patients and therapists. However, it is essential to note that these corpora were initially designed for natural language analysis and do not encompass the multimodal aspects in the accompanying videos. Thus, our study aims to fill this gap by considering and analyzing the multimodal elements within these videos, enabling a more comprehensive understanding of therapeutic interactions. These corpora initially focus on the task goal of the conversation; we perform additional annotations to study the social aspect of motivational interviewing.
\paragraph{AnnoMi}
AnnoMI is a publicly available dataset of 132 expert-annotated MI videos. Twelve were removed because they were no longer accessible (3) or of poor quality (9). The selected part of AnnoMi represents 13 hours and 25 minutes of videos. The videos are transcribed and classified between high- and low-quality MI. Each turn is annotated based on the primary therapist's verbal behavior (question, reflection, other) and the patient's talk type (neutral, change, sustain).

\paragraph{Motivational Interviewing Dataset (MID)}
The MID is similar to AnnoMI, comprising 247 publicly available simulated MI conversations. Eighty-two of these are removed because some of the videos are already part of the AnnoMI dataset (39), are missing (37), or are of poor quality (6). The selected part of the corpus represents 7 hours and 56 minutes. The videos are transcribed, but the transcriptions are of poor quality.

\paragraph{Videos pre-processing}

\begin{figure}[htp]
\centering
\begin{subfigure}{.16\linewidth}
\centering
    \includegraphics[width=\linewidth]{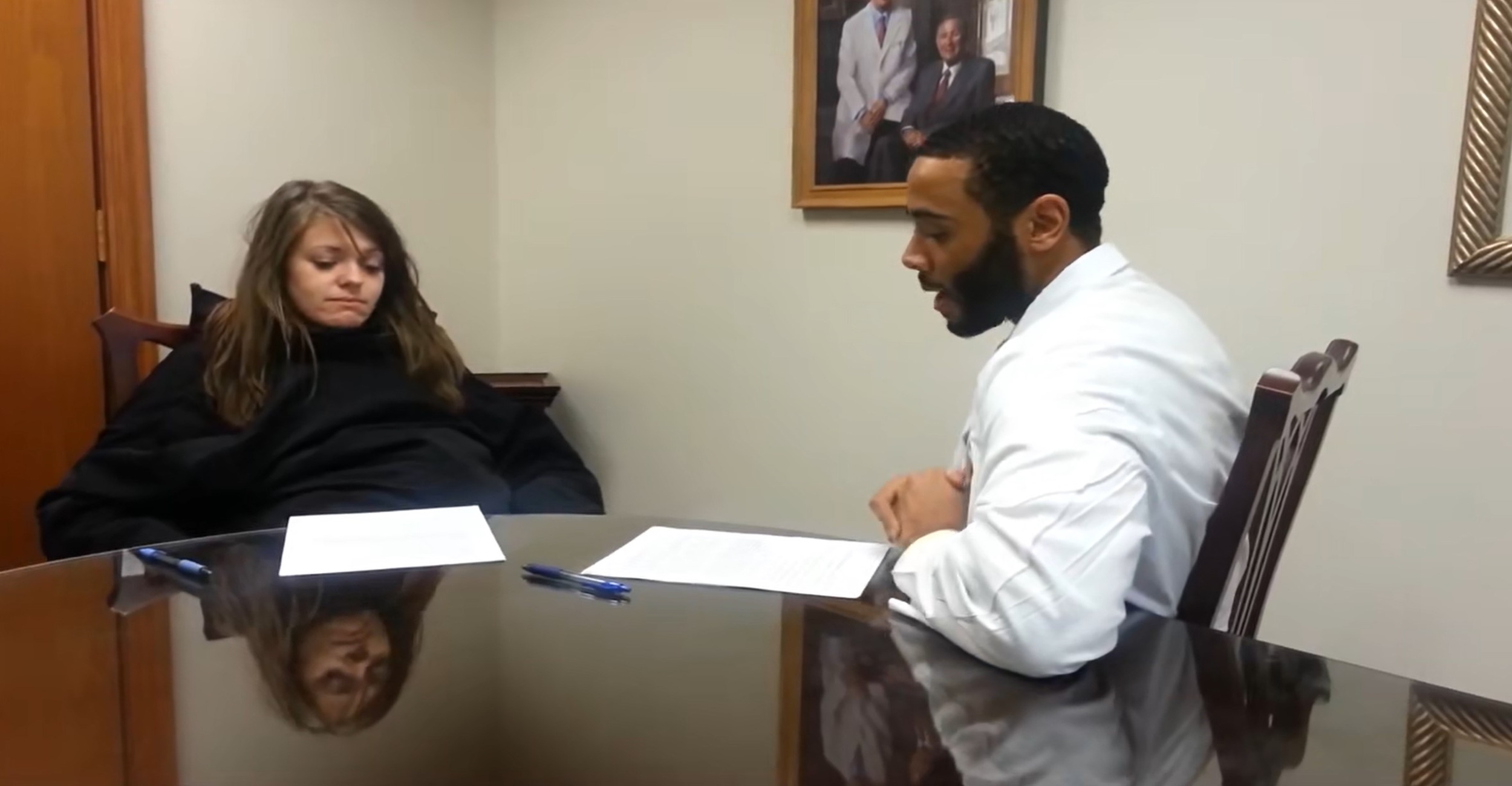}
    \caption{Both interlocutor visible}
\end{subfigure}
\begin{subfigure}{.16\linewidth}
\centering
    \includegraphics[width=\linewidth]{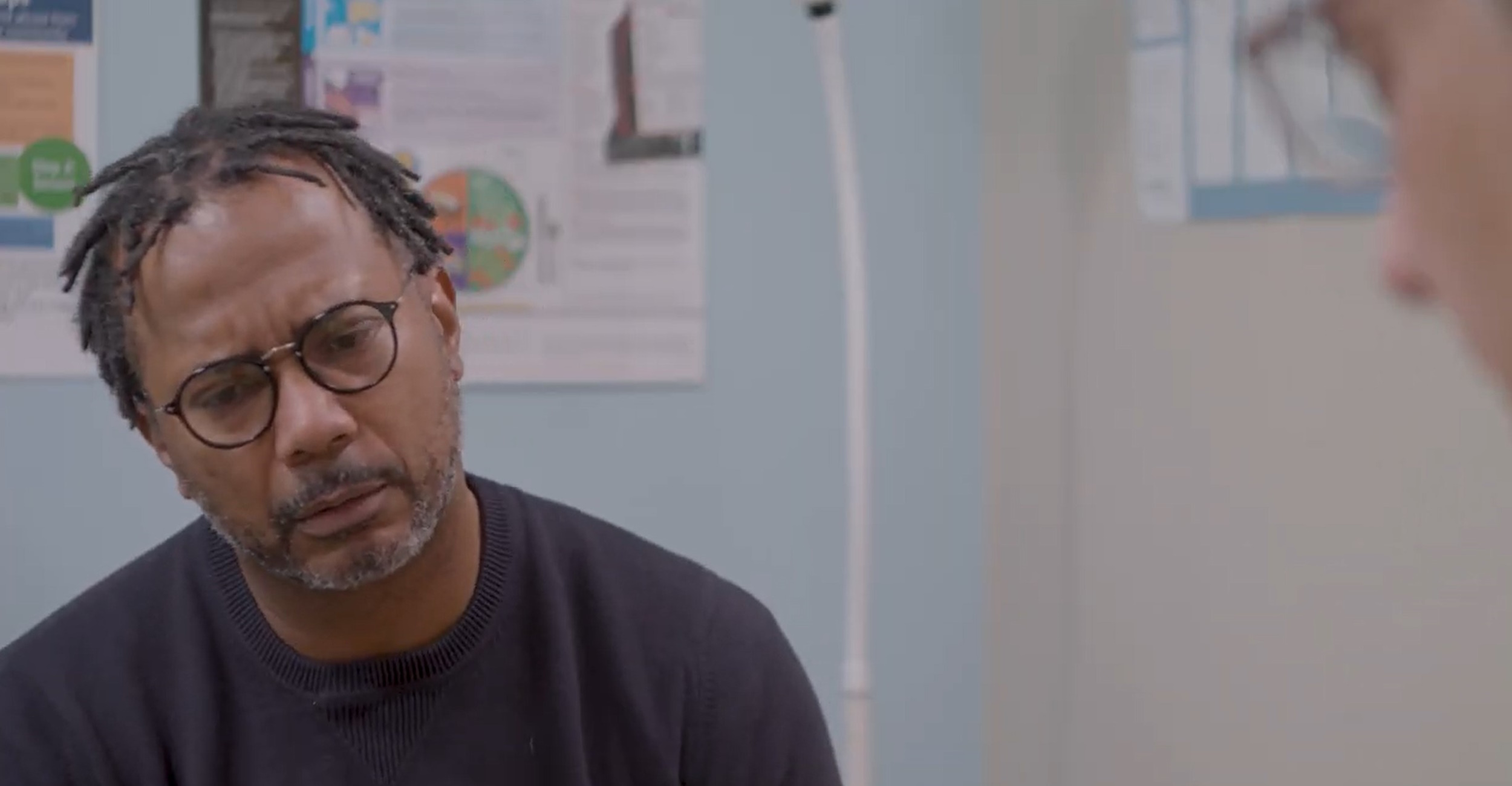}
    \caption{Only patient visible}
\end{subfigure}
\begin{subfigure}{.16\linewidth}
\centering
    \includegraphics[width=\linewidth]{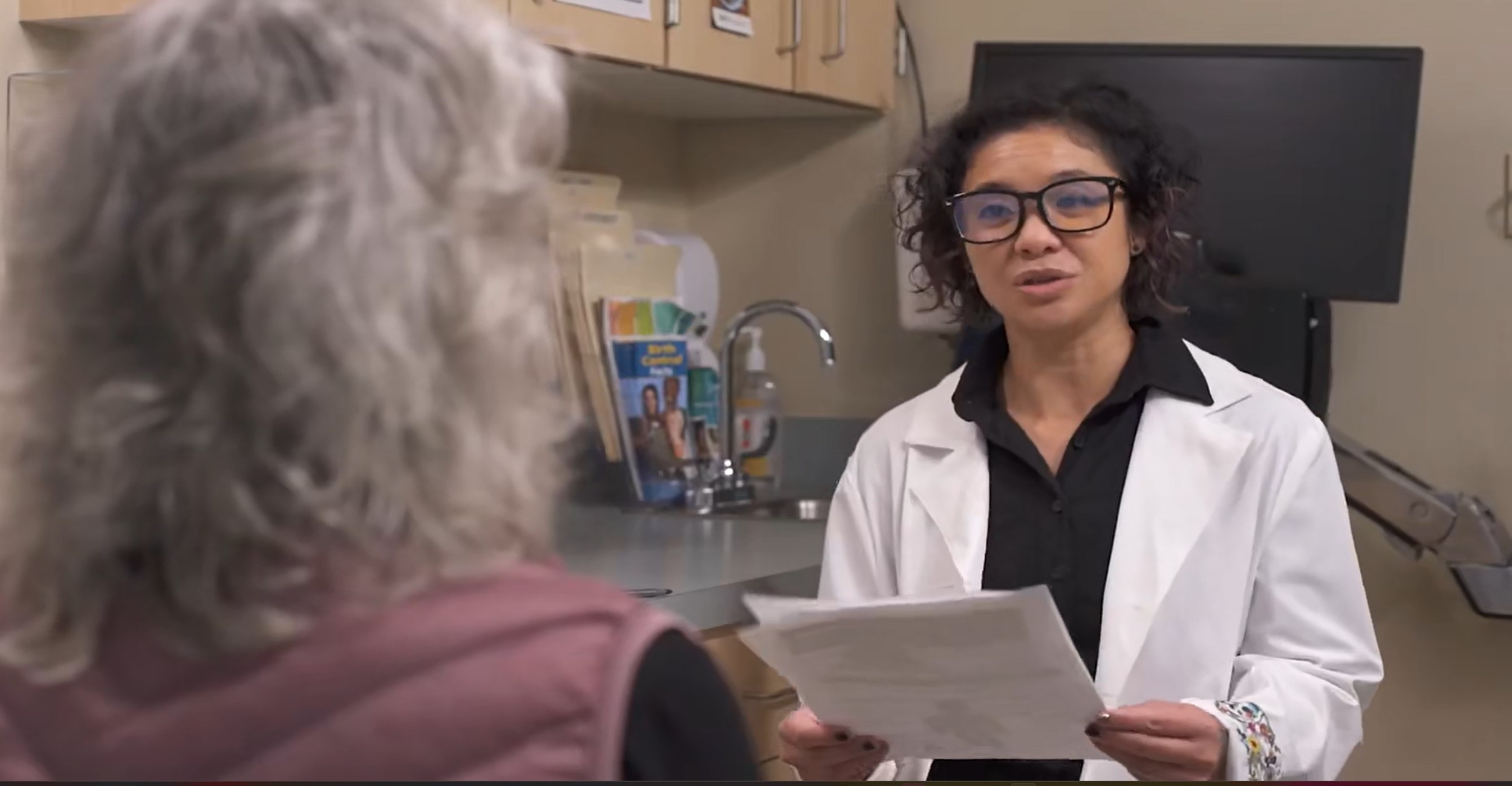}
    \caption{Only therapist visible}
\end{subfigure}
\centering
\begin{subfigure}{.16\linewidth}
    \includegraphics[width=\linewidth]{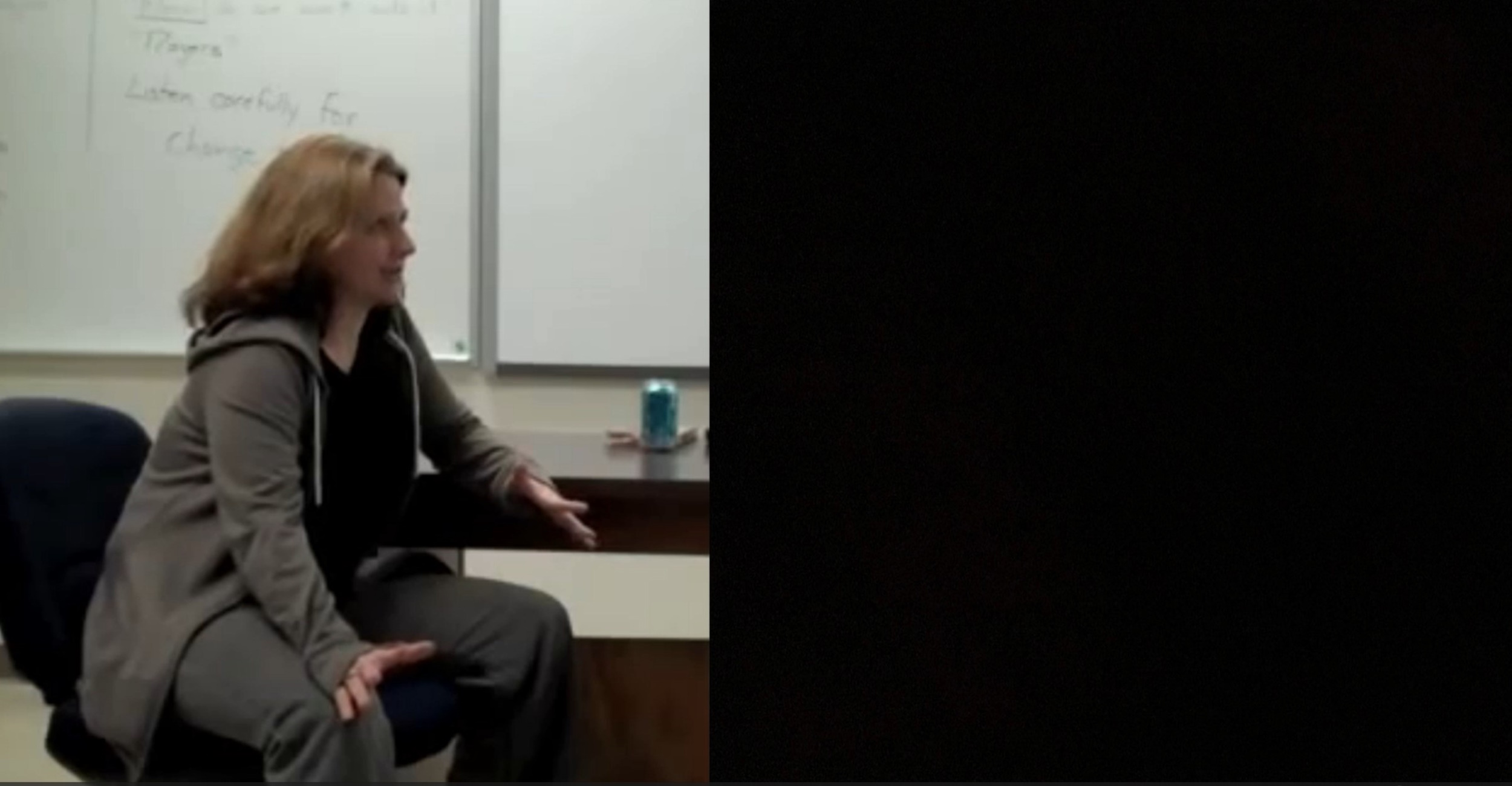}
    \caption{Edited patient visible}
\end{subfigure}
\centering
\begin{subfigure}{.16\linewidth}
    \includegraphics[width=\linewidth]{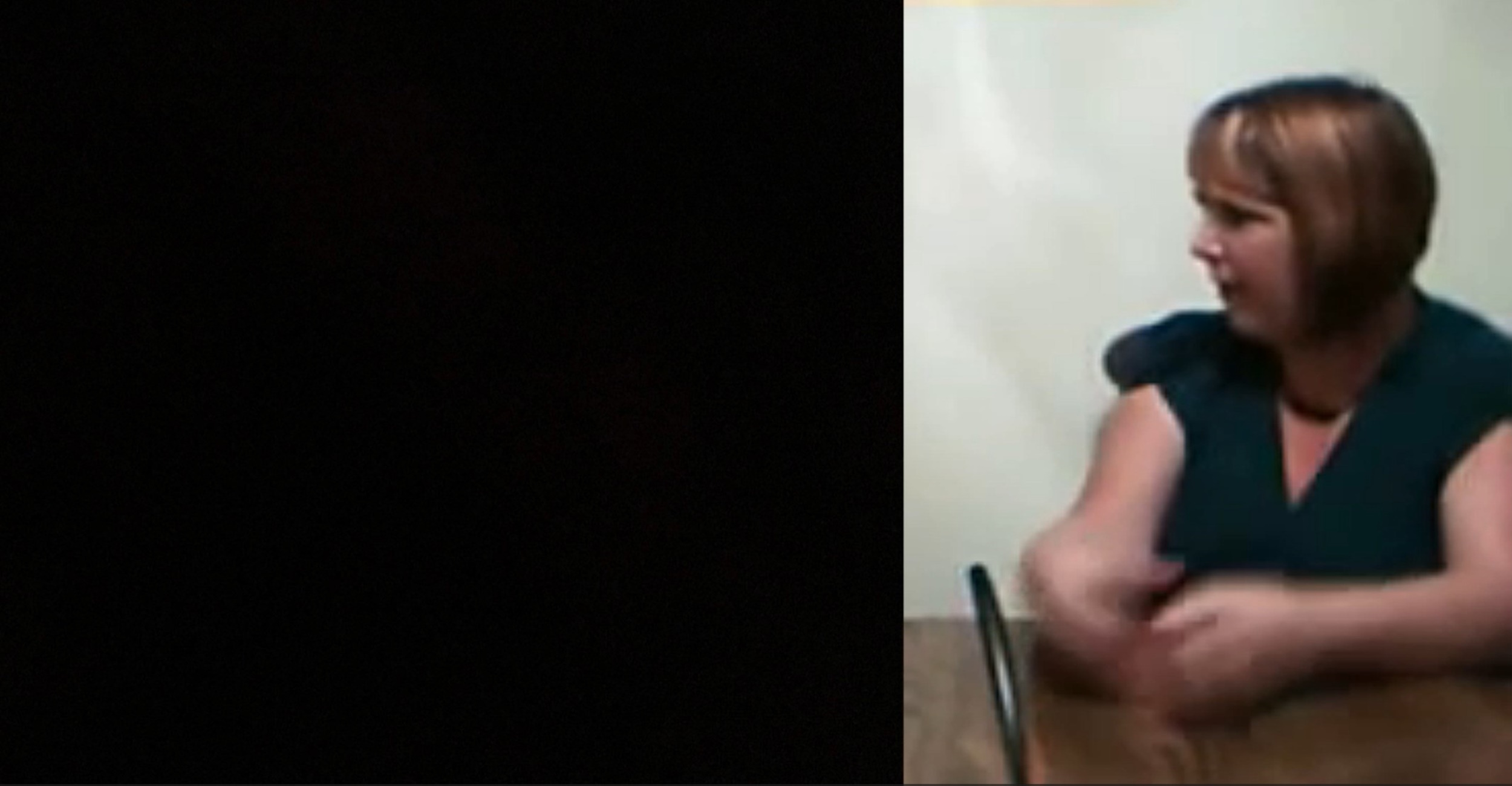}
    \caption{Edited therapist visible}
\end{subfigure}

\caption{Example of video settings}
\label{fig:example_videos}
\end{figure}
In preparation for automated annotations, the videos from the MID are transcribed anew using the transcription service \href{https://www.descript.com/transcription}{Descript}. To facilitate analysis and comparison of simultaneous nonverbal behavior, the original videos, which featured different views (i.e., patient-only, therapist-only, or both visible from the side), are edited to create three standardized views for each video: patient-only view, therapist-only view, and both patient and therapist visible (see Fig.\ref{fig:example_videos}). These different views imply that non-verbal behaviors are not accessible to every interlocutor at every instant, which should be considered for the subsequent analysis. By segregating the videos into these three distinct views, we can quickly identify and compare the visible individual's non-verbal behavior during relevant interaction segments. Furthermore, two additional annotations are generated: An annotation of who is currently visible in the video (patient, therapist, or both) and an annotation of the speaker identity (i.e., who has the speaking turn, the therapist or the patient).

\section{Multimodal annotations}
\label{sec:annot}
In the following sections, we study different aspects of the therapist and patient behaviors. We examine task-related behaviors, encompassing the dialog acts used during the conversation that reflect the classical steps of MI \cite{beauvais2019motivational}. Additionally, we explore the concept of behavior expressivity, gauged through indicators like gesture amplitude and loudness, which offer insights into the confidence levels of the interlocutors and are also a sign of empathy \cite{decker2014development}. This dimension helps us understand how self-assured the therapist and patient feel during the interaction. Our investigation also encompasses social-related behaviors, such as smiles and social dialog acts, which provide valuable insights into the evolving dynamics and rapport between the therapist and the patient \cite{jani2012role}. Finally, we look at the alignment between the therapist and the patient through verbal alignment (How much verbal expressions of the interlocutor are reused) and amplitude alignment (How similar the amplitude of the two interlocutors are)\cite{lord2015more}. 
Since interlocutors' faces are not always fully visible in the videos, we choose not to analyze facial expression alignment.
\paragraph{Separation in dialog turns}
To analyze the interaction, the corpus is separated into dialog turns. A dialog turn comprises a sequence of dialog acts that correspond to utterances. During a dialog turn, the listener can produce a backchannel that does not interrupt the turn. For each dialog turn, we give the number of backchannels performed by the listener and their timing and form. 
\paragraph{Dialog act}
\label{sec:annot_da}
% Dialog act is annotated using a combination of the Python package \href{https://pypi.org/project/DialogTag/}{dialogtag} focused on tasks. The therapist's sentences are annotated on a second dimension using the module empathic Intent \cite{rashkin2018towards} focused on empathy.
% These two annotations are combined to form a two-dimensional annotation. The categories of dialog acts are re-categorized as other if they occur less than 1\% % of the time. The final annotation categorizes each sentence into a question, a statement with opinion, a statement without opinion, appreciation, action-directive, agreement, hedging, and feedback for the task dimension, and neutral or empathic (acknowledgment, agreement, suggestion, and sympathizing) for the social dimension. This automatic dialog act annotation is validated by one human annotator manually annotating 5\% % of the corpus with Cohen Kappa of agreement with the automatic annotation of 0.52 for empathic dialog act and 0.42 for dialog acts. The many classes in the annotation schema could explain this low number. Indeed, this agreement goes up to 0.58 if we do not consider the under-represented classes of dialog acts in our test set (less than ten examples out of 1000).
Dialog acts are annotated using a schema derived from \cite{chiu2024computational}. Using multilabel annotation, patients' dialog turns are classified into one or multiple of the dialog acts defined in Table \ref{tab:da_f1_patient}. Therapists' dialog turns are classified into one or multiple of the task-oriented dialog acts and/or one of the socially oriented dialog acts described in Table \ref{tab:da_f1_therapist}. The dialog turns are classified using Mistral instruct \cite{jiang2023mistral} and few shots learning. Mistral instruct is an open-source, free model usable with an NVIDIA RTX A2000 GPU 8GB. The prompt is derived from \cite{chiu2024computational}. The automatic annotations are tested on a subpart of the HOPE dataset \cite{malhotra2022speaker}. The HOPE dataset comprises CBT and MI videos with good-quality transcripts overlapping with AnnoMI. The resulting automatic classifier will, therefore, be transferable to EMMI. We used HOPE to test the classifier and validate it on both MI and CBT.500 dialog turns of the patient and 500 of the therapist have been annotated twice with one week delay by one of the paper's authors with a Cohen kappa alpha agreement of 0.65 for the patient and  0.7 for the therapist. These annotated sentences are used to validate the automatic annotations with a macro-F1 score of 0.69 for the patient and the therapist (see F1 values for each dialog act in \ref{tab:da_f1_patient} for the patient and  \ref{tab:da_f1_therapist} for the therapist). The prompts and the annotations are available on github\footnote{\href{https://github.com/l-Galland/MI-DA-classification}{https://github.com/l-Galland/MI-DA-classification}}.
\begin{table*}[]
\begin{tabular}{|l|l|r|r|r|}
\hline
\textbf{}                                    & \multicolumn{1}{c|}{\textbf{Definition}}              & \multicolumn{1}{c|}{\textbf{macro-F1}}    & \multicolumn{1}{c|}{\textbf{Recall}} & \multicolumn{1}{c|}{\textbf{Accuracy}} \\ \hline
\begin{tabular}[c]{@{}l@{}}Changing \\ unhealthy behavior\end{tabular}         & \begin{tabular}[c]{@{}l@{}}The patient explicitly expresses their willingness to change\end{tabular}         &0,80{[}0,76;0,84{]}                      & 0,78{[}0,74;0,83{]}                 & 0,89{[}0,87;0,91{]}                   \\ \hline
\begin{tabular}[c]{@{}l@{}}Sustaining \\ unhealthy behavior\end{tabular}       & \begin{tabular}[c]{@{}l@{}}The patient explicitly expresses their unwillingness to change\end{tabular}         &0,61{[}0,55;0,67{]}                      & 0,60{[}0,55;0,66{]}                 & 0,89{[}0,87;0,91{]}                   \\ \hline
\begin{tabular}[c]{@{}l@{}}Sharing negative \\ feeling or emotion\end{tabular} & \begin{tabular}[c]{@{}l@{}}The patient shares a negative feeling or vision of the world\end{tabular}         &0,63{[}0,59;0,68{]}                      & 0,62{[}0,58;0,65{]}                 & 0,80{[}0,78;0,83{]}                   \\ \hline
\begin{tabular}[c]{@{}l@{}}Sharing positive \\ feeling or emotion\end{tabular} & \begin{tabular}[c]{@{}l@{}}The patient shares a positive feeling or vision of the world\end{tabular}         &0,65{[}0,54;0,77{]}                      & 0,80{[}0,59;0,99{]}                 & 0,98{[}0,96;0,99{]}                   \\ \hline
\begin{tabular}[c]{@{}l@{}}Realization or \\ Understanding\end{tabular}        & \begin{tabular}[c]{@{}l@{}}The patient realized or understood something about their problem\end{tabular}         &0,63{[}0,55;0,71{]}                      & 0,63{[}0,55;0,72{]}                 & 0,94{[}0,93;0,96{]}                   \\ \hline
\begin{tabular}[c]{@{}l@{}}Shapersonalonnal \\ information\end{tabular}       & \begin{tabular}[c]{@{}l@{}}The patient shares factual personal information about their situation \\or background\end{tabular}         &0,69{[}0,65;0,72{]}                       & 0,71{[}0,67;0,75{]}                 & 0,70{[}0,66;0,73{]}                   \\ \hline
Greeting or Closing                                                            & \begin{tabular}[c]{@{}l@{}}The patient opens or closes the conversation\end{tabular}         &\multicolumn{1}{l|}{0,76{[}0,65;0,85{]}} & 0,75{[}0,63;0,87{]}                 & 0,98{[}0,96;0,99{]}                   \\ \hline
Backchannel                                                                    & \begin{tabular}[c]{@{}l@{}}The patient acknowledge that they heard the last therapist's statement\end{tabular}         &\multicolumn{1}{l|}{0,65{[}0,59;0,72{]}} & 0,80{[}0,71;0,89{]}                 & 0,90{[}0,89;0,92{]}                   \\ \hline
\begin{tabular}[c]{@{}l@{}}Asking for medical \\ information\end{tabular}      & \begin{tabular}[c]{@{}l@{}}The patient ask for medical information\end{tabular}         &\multicolumn{1}{l|}{0,74{[}0,57;0,87{]}} & 0,86{[}0,59;0,99{]}                 & 0,99{[}0,98;0,99{]}                   \\ \hline
\multicolumn{2}{|l|}{\textbf{Macro average}}                                                                 & \textbf{0,69{[}0,65;0,72{]}}             & \textbf{0,73{[}0,69;0,77{]}}        & \textbf{0,90{[}0,89;0,90{]}}          \\ \hline
\end{tabular}
\caption{Classification scores of patients dialog act. The 95\% confidence intervals are computed using the bootstrap method and a 1000 runs \cite{efron1994introduction}}
\label{tab:da_f1_patient}

\end{table*}
\begin{table*}[t]
\begin{tabular}{|llrrr|}
\hline
\multicolumn{1}{|l|}{}                                              & \multicolumn{1}{c|}{\textbf{Definition}}                      & \multicolumn{1}{c|}{\textbf{Macro F1 score}}                & \multicolumn{1}{c|}{\textbf{Recall}}                        & \multicolumn{1}{c|}{\textbf{Accuracy}} \\ \hline
\multicolumn{5}{|c|}{\textbf{Task oriented Dialog acts}}                                                                                                                                                                                              \\ \hline
\multicolumn{1}{|l|}{\begin{tabular}[c]{@{}l@{}}Ask for consent \\ or validation\end{tabular}}       &          \multicolumn{1}{l|}{\begin{tabular}[c]{@{}l@{}}The therapist checks that their last statement was correct or\\ that the patient consented to move forward\end{tabular}}                                          & \multicolumn{1}{l|}{0,67{[}0,55;0,78{]}}          & \multicolumn{1}{r|}{0,77{[}0,61;0,98{]}}          & 0,97{[}0,96;0,98{]}          \\ \hline
\multicolumn{1}{|l|}{\begin{tabular}[c]{@{}l@{}}Medical \\ Education \\ and Guidance\end{tabular}}        &       \multicolumn{1}{l|}{The therapist provides the patient with medical or therapeutic facts}                                           & \multicolumn{1}{r|}{0,83{[}0,74;0,89{]}}          & \multicolumn{1}{r|}{0.83{[}0,73;0,91{]}}          & 0,97{[}0,96;0,98{]}          \\ \hline
\multicolumn{1}{|l|}{\begin{tabular}[c]{@{}l@{}}Planning with \\ the patient\end{tabular}}          &        \multicolumn{1}{l|}{\begin{tabular}[c]{@{}l@{}}The therapist builds a plan with the patient to modify\\ their unhealthy behavior/thoughts\end{tabular}}                                             & \multicolumn{1}{r|}{0,73{[}0,67;0,79{]}}          & \multicolumn{1}{r|}{0,70{[}0,65;0,76{]}}          & 0,92{[}0,90;0,94{]}          \\ \hline
\multicolumn{1}{|l|}{Give Solution}                                &                  \multicolumn{1}{l|}{The therapist provides the patient with solutions to solve their problem}                         & \multicolumn{1}{r|}{0,66{[}0,59;0,73{]}}          & \multicolumn{1}{r|}{0.67{[}0,59;0,76{]}}          & 0,94{[}0,92;0,95{]}          \\ \hline
\multicolumn{1}{|l|}{\begin{tabular}[c]{@{}l@{}}Ask about \\current \\ emotions\end{tabular}}       &                 \multicolumn{1}{l|}{The therapist ask the patient what they are feeling during the therapy session}                                      & \multicolumn{1}{r|}{0,64{[}0,60;0,69{]}}          & \multicolumn{1}{r|}{0,62{[}0,57;0,66{]}}          & 0,88{[}0,85;0,90{]}          \\ \hline
\multicolumn{1}{|l|}{\begin{tabular}[c]{@{}l@{}}Invite to shift \\ outlook\end{tabular}}        &              \multicolumn{1}{l|}{\begin{tabular}[c]{@{}l@{}}The therapist asks the patient to imagine their reaction to a future event or\\ to change their perspectives on a past even\end{tabular}}                                           & \multicolumn{1}{r|}{0.57{[}0.51;0.63{]}}          & \multicolumn{1}{r|}{0.56{[}0,52;0,61{]}}          & 0,91{[}0,88;0,93{]}          \\ \hline
\multicolumn{1}{|l|}{\begin{tabular}[c]{@{}l@{}}Ask for \\Information\end{tabular}}              &                 \multicolumn{1}{l|}{\begin{tabular}[c]{@{}l@{}}The therapist asks the patient factual information \\ about their background or situation\end{tabular}}                                      & \multicolumn{1}{r|}{0.78{[}0.75;0.81{]}}          & \multicolumn{1}{r|}{0,81{[}0,78;0,84{]}}          & 0,80{[}0,77;0,83{]}          \\ \hline
\multicolumn{1}{|l|}{Reflection}                                                &         \multicolumn{1}{l|}{The therapist summarize or reformulate the patient statement without judgment}                     & \multicolumn{1}{r|}{0.73{[}0.69;0.76{]}}          & \multicolumn{1}{r|}{0.72{[}0,68;0,76{]}}          & 0,77{[}0,74;0,80{]}          \\ \hline
\multicolumn{5}{|c|}{\textbf{Socially oriented Dialog acts}}                                                                                                                                                                                          \\ \hline
\multicolumn{1}{|l|}{\begin{tabular}[c]{@{}l@{}}Empathic \\ reaction\end{tabular}}          &               \multicolumn{1}{l|}{The therapist expresses empathy to the patient}                                              & \multicolumn{1}{l|}{0,62{[}0,53;0,71{]}}          & \multicolumn{1}{r|}{0.68{[}0,55;0,82{]}}          & 0,96{[}0,94;0,97{]}          \\ \hline
\multicolumn{1}{|l|}{\begin{tabular}[c]{@{}l@{}}Acknowledge \\ progress and \\ encourage\end{tabular}} &\multicolumn{1}{l|}{The therapist praises the patient for their achievements or encourages them}  & \multicolumn{1}{l|}{0,62{[}0,53;0,71{]}}          & \multicolumn{1}{r|}{0.87{[}0,67;0,98{]}}          & 0,96{[}0,94;0,97{]}          \\ \hline
\multicolumn{1}{|l|}{Backchannel}                                                    &       \multicolumn{1}{l|}{The therapist acknowledges that they heard the last patient's statement}                  & \multicolumn{1}{l|}{0,65{[}0,56;0,74{]}}          & \multicolumn{1}{r|}{0,83{[}0,68;0,98{]}}          & 0,96{[}0,95;0,97{]}          \\ \hline
\multicolumn{1}{|l|}{\begin{tabular}[c]{@{}l@{}}Greeting or \\Closing\end{tabular}}         &           \multicolumn{1}{l|}{The therapist open or closes the conversation}                                                 & \multicolumn{1}{l|}{0,85{[}0,77;0,93{]}}          & \multicolumn{1}{r|}{0.88{[}0,79;0,96{]}}          & 0,98{[}0,96;0,95{]}          \\ \hline
\multicolumn{1}{|l|}{\begin{tabular}[c]{@{}l@{}}Experience \\Normalization \\ and Reassurance\end{tabular}}  &\multicolumn{1}{l|}{The therapist normalizes the patient experience and reassure them}   & \multicolumn{1}{r|}{0,59{[}0,49;0,68{]}}          & \multicolumn{1}{r|}{0.58{[}0,49;0,68{]}}          & 0,97{[}0,95;0,98{]}          \\ \hline
\multicolumn{2}{|l|}{\textbf{Macro average}}                                                & \multicolumn{1}{r|}{\textbf{0,69{[}0,67;0,71{]}}} & \multicolumn{1}{r|}{\textbf{0.73{[}0,70;0,76{]}}} & \textbf{0.92{[}0,91;0,93{]}} \\ \hline
\end{tabular}
\caption{Classification scores of therapists dialog act. The 95\% confidence intervals are computed using the bootstrap method and a 1000 runs \cite{efron1994introduction}}
\label{tab:da_f1_therapist}
\end{table*}
\paragraph{Verbal alignment} is computed using Dialign \cite{dubuisson2021towards}.
It computes the number of reused verbal expressions initiated by the other interlocutor.

\paragraph{Loudness}, the intensity of the voice is annotated using OpenSmile \cite{eyben2010opensmile}.

\paragraph{Face Features} for the therapist and the patient are collected using OpenFace \cite{baltrusaitis2018openface}. The action units are filtered with a median filter of size 3 and interpolated for missing values.
\paragraph{Smile} are detected using 
 the \href{https://github.com/srauzy/HMAD}{HMAD project} \cite{rauzy2018automatic} that annotates smiles on the five levels of the Smiling Intensity Scale based on features extracted with OpenFace.

 \paragraph{The body joints' position and orientation} are collected using OpenPose. \cite{cao2017realtime}

\paragraph{Amplitude}
The amplitude of the upper body is defined as the bounding box around the speaker for a given time frame. It is computed by dividing the distance between the two wrists by the height of the bust in the current frame.

\paragraph{Change talk}
The Motivational Interviewing Skill Code (MISC) categorizes patient talk into three distinct classifications:
\textbf{Change Talk} that denotes the expression of behaviors aligned with embracing change, \textbf{Sustain Talk} that refers to the articulation of behaviors demonstrating a rejection of change,\textbf{ Neutral Talk} that encompasses behaviors unrelated to the targeted change. The MISC classification scheme carries significance due to its demonstrated correlation with therapy outcomes \cite{magill2014technical,magill2018meta}. Moreover, the talk type has already been annotated within AnnoMI. Previous work trained a multimodal classifier using this annotated data, yielding an F1 score of 0.8 \cite{galland2023seeing}. 
This classifier is employed to annotate the MID in terms of change talk.\footnote{This classification into change/ sustain/ neutral is different from the one carried out in Section\ref{sec:annot_da} as it considers all modalities and not only the text. This classification also takes into account subtext that is not taken into account by the dialog act classifier and is therefore complementary}

\paragraph{Patient clustering}

\begin{figure}[ht]
     
        \centering
     \begin{subfigure}[b]{0.49\linewidth}
         \centering
         \includegraphics[width=\linewidth]{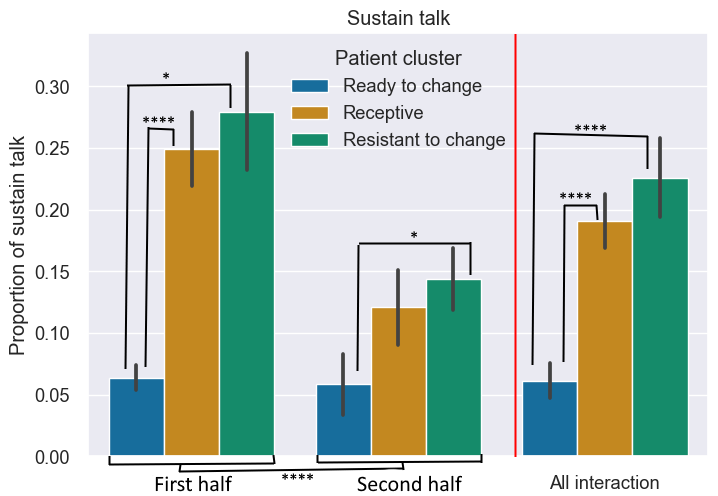}
         \caption{Sustain talk}
         \label{fig:types_sustain}
     \end{subfigure}
     \hfill
     \begin{subfigure}[b]{0.49\linewidth}
         \centering
         \includegraphics[width=\linewidth]{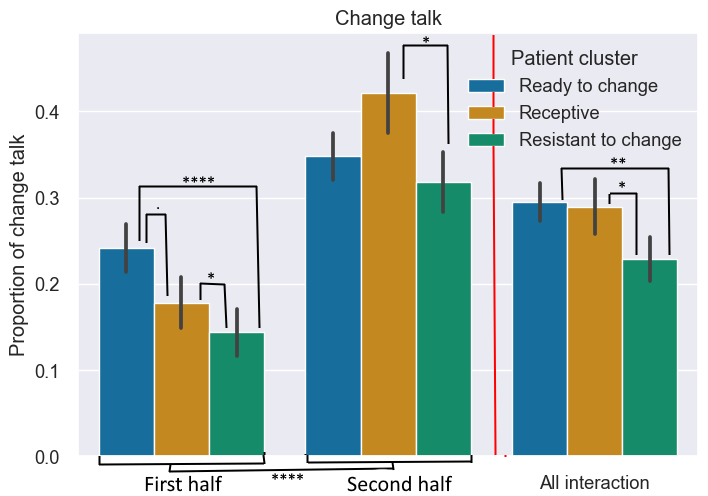}
         \caption{Change talk}
         \label{fig:types_change}
     \end{subfigure}
     \hfill

       \caption{Evolution of change and sustain talk in time according to patient type \small{. p$<$0.1, * p$<$0.05, ** p$<$0.01, ** p$<$0.001,*** p$<$0.0001, **** p$<$0.00001}}
        \label{fig:timepatienttype}
\end{figure}

Each patient utterance in EMMI is annotated at the level of change talk: [neutral, sustain, change].
This annotation is transformed into a numerical score of change talk where sustain = -1, neutral = 0, and change = 1. We apply the KMeans algorithm with a DTW metric \cite{sakoe1978dynamic} to compute clusters of patients. The DTW metric accounts for similar patterns in change talk behavior even when the timing or length of the pattern is not precisely identical. The Elbow method \cite{thorndike1953belongs} shows that the optimal number of clusters is 3 with a silhouette score of 0.17. The visualization of the clusters of patients as well as a time-based analysis (see Fig. \ref{fig:timepatienttype}) shows three types that can be concretely interpreted; Patient type (Kruskal-Wallis H test: $H_{change}(2) = 18.1$,$p_{change} = 1.2e-4$, $H_{sustain}(2) = 20.5$, $p_{sustain} = 3.5e-5$) and Time (Kruskal-Wallis H test: $H_{change}(1) = 40.7$,$p_{change} = 1.8e-10$, $H_{sustain}(1) = 12.17$, $p_{sustain} = 5.1e-4$) have a significant main effect on both change and sustain talk proportion for at least two groups. The following Mann-Whitney U post hoc test results can be seen in Table \ref{tab:type} and Figure \ref{fig:timepatienttype}. Since \cite{abuse2019enhancing} states that ``Patient stuck in ambivalence engages in a lot of sustain talk, whereas patients who are more ready to change engage in more change talk with stronger statements supporting change,'' we can then interpret our clusters as the following:
\begin{itemize}
    \item When the patient begins the conversation with significantly less sustain talk than other patients ($p<1e-4$) and significantly more change talk than other patients ($p<5e-2$), we define them as \textbf{"Ready to change"}

    \item When the patient displays significantly less change talk throughout the conversation than other patients ($p<5e-3$), we define them as \textbf{"Resistant to change"}
    \item When the patient starts with a comparable number of sustain talk as ``Resistant to change'' patients but more change talk ($p<5e-2$) and the patient displays significantly more change talk than ``Resistant to change'' patient, especially during the second half of the conversation ($p<1e-2$), we define them as \textbf{"Receptive"}
\end{itemize}
\begin{table}[ht]
\centering

\begin{tabular}{lll|l|l|}
\cline{4-5}
                                           &                          &     & $p$ change     & $p$ sustain    \\ \cline{2-5} 
\multicolumn{1}{l|}{}                      & \multicolumn{1}{l|}{Beg} & End & 4.0e-12 (****)& 3.9e-5 (****) \\ \cline{2-5}\\  \cline{2-5}
\multicolumn{1}{l|}{}                      & \multicolumn{1}{l|}{Rec} & Op  & 9.5e-1 & 5.2e-7 (****) \\ \cline{2-5} 
\multicolumn{1}{l|}{}                      & \multicolumn{1}{l|}{Rec} & Res & 3.8e-3 (**)& 2.6e-5 (****) \\ \cline{2-5}
\multicolumn{1}{l|}{}                      & \multicolumn{1}{l|}{Op}  & Res & 2.0e-2 (*)& 5.5e-1        \\  \cline{2-5} \\ \hline
\multicolumn{1}{|l|}{\multirow{3}{*}{Beg}} & \multicolumn{1}{l|}{Rec} & Op  & 9.7e-2 (.)& 1.2e-7 (****) \\ \cline{2-5} 
\multicolumn{1}{|l|}{}                     & \multicolumn{1}{l|}{Rec} & Res & 1.2e-5 (****)& 1.2e-2 (*)    \\ \cline{2-5} 
\multicolumn{1}{|l|}{}                     & \multicolumn{1}{l|}{Op}  & Res & 2.6e-2 (*)& 1.9e-1        \\ \hline \\ \hline
\multicolumn{1}{|c|}{\multirow{3}{*}{End}} & \multicolumn{1}{l|}{Rec} & Op  & 2.6e-1& 1.6e-1        \\ \cline{2-5} 
\multicolumn{1}{|c|}{}                     & \multicolumn{1}{l|}{Rec} & Res & 1.1e-1& 1.3e-2        \\ \cline{2-5} 
\multicolumn{1}{|c|}{}                     & \multicolumn{1}{l|}{Op}  & Res & 4.3e-2 (*)& 6.1e-1        \\ \hline
\end{tabular}

\caption{Post hoc test on patient talk type depending on patients type and time \small{. p$<$0.1, * p$<$0.05, ** p$<$0.01, ** p$<$0.001, *** p$<$0.0001, **** p$<$0.00001}\\ \small{Beg = First half, End=Second half, Op = Ready to change, Rec=Receptive, Res = Resistant to change}}
\label{tab:type}
\end{table}

\section{Multimodal behavior analysis}
\label{sec:analysis}
In this section, we investigate our three research questions through the length of the targeted behaviors described in Section \ref{sec:annot}: task behavior, expressivity, social behavior, and alignment. Section \ref{sec:discussion} discusses the reported statistical results. We study the differences in target behaviors between the first and second half of the conversation to study the evolution during the interaction (RQ1). We separated the interaction into two parts, as the small number of videos does not allow for more fine-grained studies. We also study the differences between different types of patient talk (RQ2) and talk types (RQ3).
We use non-parametric tests whenever the normality requirements for ANOVA are not met. All p-values are corrected using the Bonferroni method.
For clarity purposes, we do not report every test value in the following section. We also focus on the high-quality part of the corpus as we aim to develop a high-quality virtual therapist.

\subsection{Task related behavior}

\paragraph{Therapist dialog acts}
Therapist plan more with the patient during the second half than during the first half of the conversation (T-test: $p=0.02$). There is also a difference in the use of planning according to at least two groups of patient types (Kruskal-Wallis H test: $p = 0.03$). Therapists plan significantly more with ``Receptive'' than with ``Resistant to change'' (Mann-Whitney U test: $p=0.04$). The therapist asks for more information during the first half of the conversation (T-test: $p=0.003$), but there are no significant differences between patients' talk types (Kruskal-Wallis H test: $p=0.84$). There are no significant differences between the first and the second half of the conversation in the use of invitation to shift outlook (T-test: $p=0.12$). However, there is a difference between at least two groups of patients for the invitation to shift outlook (Kruskal-Wallis H test: $p = 0.02$). The therapist invites significantly more to shift outlook when interacting with a ``Receptive'' patient than when interacting with a ``Resistant to change'' patient (Mann-Whitney U test: $p=0.05$).

% The therapist uses more statements with opinions during the second half than during the first half of the conversation(T-test: $p=3e-6$). There is also a difference in the use of opinionated statements according to at least two groups of patient types (Kruskal-Wallis H test: $p = 0.003$). Therapists use significantly fewer opinionated statements when the patient is ``Resistant to change'' than when the patient is ``Receptive'' (Mann-Whitney U test: $p=0.003$).The therapist asks more questions during the first half of the conversation (T-test: $p=1e-5$), but there are no significant differences in the patient's talk type (Kruskal-Wallis H test: $p=0.8$). Statement without opinion, appreciation, action-directive, agreement, hedging, and feedback are used consistently across the conversation (T-test between first and second half: all $p>0.2$) and patient's types (Kruskal-Wallis H test: all $p>0.2$).
 
\paragraph{Patient dialog act}
The patient expresses significantly more intention to change unhealthy behaviors during the second half of the conversation than during the second half (T-test: $p=0.0005$). There is also a difference in the use of changing statements according to at least two groups of patient types (Kruskal-Wallis H test: $p = 0.03$). Patients use significantly fewer change statements when they are ``Resistant to change'' than when they are ``Receptive'' (Mann-Whitney U test: $p=0.03$). Patients share more information during the first half of the conversation than during the second half (T-test: $p = 0.002$). Yet, there are no differences between patient types (Kruskal-Wallis H test $p=0.73$).  

\subsection{Expressivity related behaviors}
\paragraph{Amplitude}
Patients show significantly more amplitude during the first half than during the second half of the conversation (T-test: $p = 0.04$), while no significant differences are observed for the amplitude of the therapist between the first and the second half of the conversation (T-test: $p=0.45$); significant differences are observable between patient types for the patient's body amplitude (Kruskal-Wallis H test: $p= 7.4e-11$). ``Resistant to change'' patients display more amplitude, while ``Receptive'' patients display smaller motions with less amplitude (Mann-Whitney U test: all $p <1e-4$). There are no significant variations in amplitude between groups for patients (Kruskal-Wallis H test: $H(2) = 0.06$, $p = 0.97$) and therapists (Kruskal-Wallis H test: $H(2) = 3.7$, $p = 0.16$).

% \begin{figure}[ht]
%      \centering
%      \begin{subfigure}[b]{0.45\linewidth}
%          \centering
%          \includegraphics[width=\linewidth]{figures/openness_client.png}
%          \caption{Patient amplitude}
%          \label{fig:openness_patient}
%      \end{subfigure}
%      \hfill
%      \begin{subfigure}[b]{0.45\linewidth}
%          \centering
%          \includegraphics[width=\linewidth]{figures/therapist_openness.png}
%          \caption{Therapist amplitude}
%          \label{fig:openness_therapist}
%      \end{subfigure}
%      \hfill
     
%         \caption{Evolution of amplitude separated by patient type \small{. p<$0$.1, * p$<$0.05, ** p$<$0.01, ** p$<$0.001, *** p$<$0.0001, **** p$<$0.00001}}
%         \label{fig:openness}
% \end{figure}

\begin{figure}[ht]
     \centering
     \begin{subfigure}[b]{0.49\linewidth}
         \centering
         \includegraphics[width=\linewidth]{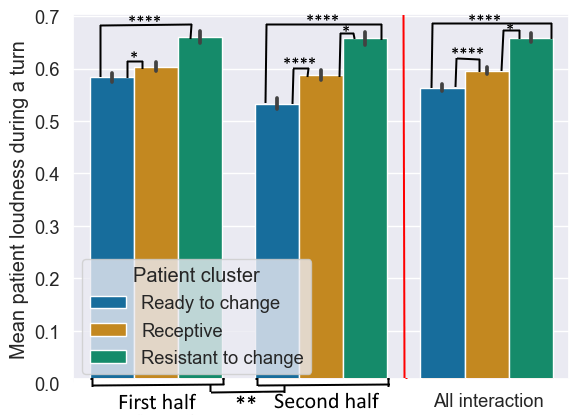}
         \caption{Patient loudness}
         \label{fig:loudness_patient}
     \end{subfigure}
     \hfill
     \begin{subfigure}[b]{0.49\linewidth}
         \centering
         \includegraphics[width=\linewidth]{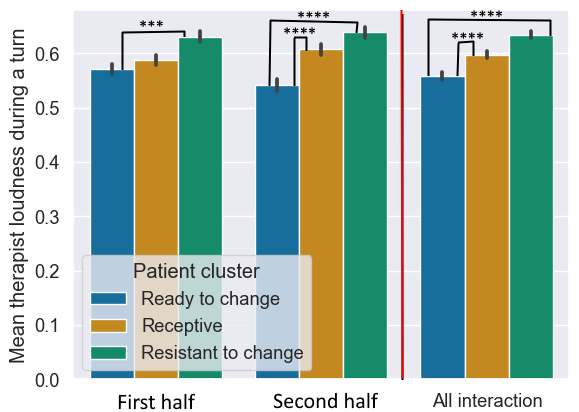}
         \caption{Therapist loudness}
         \label{fig:loudness_therapist}
     \end{subfigure}
     \hfill
   
        \caption{Evolution of loudness separated by types \small{. p$<$0.1, * p$<$0.05, ** p$<$0.01, ** p$<$0.001,*** p$<$0.0001, **** p$<$0.00001}}
        \label{fig:loudness}
\end{figure}

\paragraph{Loudness}
There are variations in patient loudness between the start of the conversation and its ending (T-test: $p = 0.003$). Specifically, patients exhibit a higher voice volume towards the dialogue's end, although such a trend isn't mirrored in therapist loudness (T-test: $p = 0.29$). There are notable differences in loudness across various patient types for both patients and therapists (Kruskal-Wallis H test: $H_{patient}(2) = 53.18$, $p_{patient} = 2.84e-12$ and $H_{therapist}(2) = 58.9$, $p_{therapist} = 1.61e-13$). Both patients and therapists speak more loudly when they resist change and adopt a softer tone when the patient is ``Ready to change'' (Mann-Whitney U test: all $p < 5e-2$). This difference is not significant between the therapists interacting with ``Receptive'' and ``Resistant to change'' patients (Mann-Whitney U test:$p=0.3$) (see Fig.\ref{fig:loudness}). Furthermore, loudness discrepancies related to expressed talk types emerge among patients (Kruskal-Wallis H test: $H(2) = 23.47$, $p = 8e-6$). Patients speak at a higher volume when expressing ``sustain talk'' than ``change talk'' and ``neutral talk'' (Mann-Whitney U test: all $p <0.006$).

% \begin{figure}[ht]
%      \centering
%      \begin{subfigure}[b]{0.49\linewidth}
%          \centering
%          \includegraphics[width=\linewidth]{figures/loudness_patient_talk_type.png}
      
%          \caption{Patient loudness}
%          \label{fig:loudness_talk_patient}
%      \end{subfigure}
%      \hfill
%      \begin{subfigure}[b]{0.49\linewidth}
%          \centering
%          \includegraphics[width=\linewidth]{figures/therapist_loudness_talk_type.png}
    
%          \caption{Therapist loudness}
%          \label{fig:loudness_talk_therapist}
%      \end{subfigure}
%      \hfill
          
%         \caption{Evolution of loudness separated by talk types \small{. p$<$0.1, * p$<$0.05, ** p$<$0.01, ** p$<$0.001,*** p$<$0.0001, **** p$<$0.00001}}
%         \label{fig:loudness_talk}
% \end{figure}
\subsection{Social related behaviors}

\paragraph{Therapist social dialog acts}
% The therapist uses significantly more empathic dialog during the second half of the conversation than in the first half (T-test: $p=0.004$). Therapists use more suggestions (T-test: $p=0.0003$) during the second half of the conversation. There are no differences in the therapist's number of empathic dialog acts depending on the patient type (Kruskal-Wallis H test: $p=0.6$).
There are no significant differences between the first and the second half of the conversation in the use of empathic reaction (T-test: $p=0.41$). However, there is a difference between at least two groups of patients (Kruskal-Wallis H test: $p = 0.04$). The therapist tends to use more empathic reactions when interacting with a ``Receptive'' patient than when interacting with a ``Resistant to change'' patient (Mann-Whitney U test: $p=0.08$).

\paragraph{Smiles}
There are no significant differences in the number of therapists' smiles between the first and second half of the conversation (T-test: $p=0.4$). However, their smiles last longer in the second half (T-test: $p<1e-6$). On the other hand, the patient smiles less during the second half of the interaction (T-test: $p<1e-46$) but with more prolonged (T-test: $p<1e-15$) and more intense smiles (T-test: $p<1e-22$). There is a difference between at least two groups of patients in smiles of patients and therapists for the number of smiles (Kruskal-Wallis H test: $p_{patient}<1e-30$, $p_{therapist}<1e-30$), duration (Kruskal-Wallis H test: $p_{patient}<1e-19$, $p_{therapist}<1e-12$) and intensity (Kruskal-Wallis H test: $p_{patient}<1e-10$, $p_{therapist}<1e-6$). Patients smile more when they are ``Resistant to change'' and less when they are ``Ready to change'' (Mann-Whitney U test: all $p <1e-30$). However, patients' smiles are longer when they are ``Ready to change'' and less long when they are ``Resistant to change'' (Mann-Whitney U test: all $p < 1e-2$). Smiles are also significantly more intense when patients are ``Resistant to change'' and less intense when they are ``Ready to change'' (Mann-Whitney U test: all $p<1e-5$). Therapists smile more when patients they are interviewing are ``Resistant to change'' and less when they are ``Ready to change'' (Mann-Whitney U test: all $p <1e-30$). Smiles last longer when the patient is ``Ready to change'' and shorter when they are ``Receptive'' (Mann-Whitney U test: all $p < 5e-2$). Smiles are more intense when the patient is ``Resistant to change'' and less intense when the patient is ``Ready to change'' (Mann-Whitney U test: all $p<1e-30$). There is a difference between at least two different types of patient talk for the therapists' and the patients' smiles in terms of number (Kruskal-Wallis H test: $p_{patient}<7e-3$, $p_{therapist}<1e-27$), duration (Kruskal-Wallis H test: $p_{patient}<1e-30$, $p_{therapist}<1e-30$) and intensity (Kruskal-Wallis H test: $p_{patient}<4e-6$, $p_{therapist}<1e-11$). Patients smile more when they express neutral talk than when they listen (Mann-Whitney U test: $p<5e-2$). When the patient is listening, their smiles last significantly longer (Mann-Whitney U test: $p<1e-30$), and their smiles are significantly more intense than when they have the speaking turn and express neutral or sustain talk (Mann-Whitney U test: $p<3e-2$). Therapists smile significantly more when they have the turn (Mann-Whitney U test: $p < 1e-7$); they also smile more when they listen to neutral talk than when they listen to change talk (Mann-Whitney U test: $p=3e-2$). They also smile significantly longer(Mann-Whitney U test: $p<1e-30$) and intensely (Mann-Whitney U test: $p<9e-5$) when they have the turn.

\subsection{Alignment}
\paragraph{Verbal alignment}
The Dialign tools \cite{dubuisson2021towards} provide for each utterance the list of verbal expressions initiated by an interlocutor that the other interlocutor reuses.
Throughout the dialogue, there is a discernible upward trend in the recurrence of verbal expressions, both for therapist-initiated and patient-initiated utterances. There is a notable variance in repeated expression based on the type of patient talk and the type of patient (Kruskal-Wallis H test: all $p < 0.02$). A higher frequency of expressions is reused when patients convey neutral statements (Mann-Whitney U test: all $p< 1e-11$). Additionally, the therapist reuses more of the patient's expressions when the patient is categorized as ``Ready to change'' or ``Receptive'' (Mann-Whitney U test: all $p < 3e-4$).

\paragraph{Amplitude alignment}
The alignment of the amplitude between the patients and the therapists is quantified by the absolute value of the difference between the therapist's and the patient's amplitude. There are statistical differences in alignment between at least two types of patients (Kruskal-Wallis H test: $H(2) = 12$, $p =0.002$). ``Receptive'' patients are significantly more aligned with their therapists than ``Resistant to change'' patients (Mann-Whitney U test: $p=1e-6$) and ``Ready to change'' patients (Mann-Whitney U test: $p=0.02$). There are no differences between patient talk types (Kruskal-Wallis H test: $H(2) = 0.19$, $p < 0.9$).

\section{Discussion}
\label{sec:discussion}
\paragraph{Task related}
Throughout the conversation, we observe an evolution of the dialog acts used by the therapist and the patient \textbf{(RQ1)}. During the first half of the conversation, the therapist asks many questions to assess the context and concerns of the patient. In contrast, the patient shares personal information to answer those questions. During the second half of the conversation, the therapist makes more statements about planning and advises the patients on addressing their problems. In response, the patient has more intention to change, reacting to the therapist's input. However, this global evolution can vary depending on the type of patient \textbf{(RQ2)}. The therapist uses more invitation to shift outlook when the patient is classified as ``Receptive'' than when they are classified as ``Resistant to change.'' This may imply that therapists attempt to provide new perspectives to patients who appear open to receiving them but have not yet determined their motivation to change (namely, the ``Receptive'' ones). Furthermore, patients classified as ``Resistant to change'' tend to manifest behaviors indicative of a reluctance to engage in dialogue and use significantly fewer statements explicitly directed toward changing their behaviors.
\paragraph{Behavior expressivity}
We observe an evolution in the behavior expressivity of the patients throughout the conversation \textbf{(RQ1)}. Indeed, patients tend to be physically more expressive in the second half of the conversation, where they talk more loudly and display behavior with larger amplitude. These observable changes suggest a growing confidence and self-assurance as the dialogue unfolds. Differences in this degree of expressivity are also observable between patient types \textbf{ (RQ2)}. In particular, ``Resistant to change'' patients exhibit a more pronounced behavior expressivity, characterized by louder speech and a more openly engaged demeanor than other patient types. This observation implies that these patients are significantly more assertive and self-assured about their viewpoints \cite{vickers2010loudness}. Similarly, patients are louder when they perform sustain talk \textbf{(RQ3)}. An interpretation of these results is that patients who are ``Resistant to change'' or expressing sustained talk when faced with an MI therapist are confident in their beliefs \cite{scherer1973voice}, which might explain why it is more difficult to change their minds.

\paragraph{Social behaviors}
The interpersonal relations between the patient and the therapist improve significantly as the interaction progresses \textbf{(RQ1)}. The duration and intensity of smiles from the patient and the therapist increase in the second half of the interaction. Social behaviors are particularly pronounced when the patient engages in what can be termed ``neutral talk'' \textbf{(RQ3)}, characterized by a more positive outlook and longer-lasting smiles. Interestingly, the patient also tends to convey a more positive demeanor when expressing ``change talk'' instead of ``sustain talk''.
Furthermore, when both the therapist and the patient express shared goals, there is a distinct increase in social behaviors between the two interlocutors. While it is noted that social behaviors are generally more present with patients who are ``Ready to change'' compared to those who exhibit resistance \textbf{(RQ2)}, this relationship is nuanced. Therapists tend to use more emphatic statements with Receptive patients than patients who are ``Resistant to change''. ``Ready to change'' patients display a more positive disposition, with longer-lasting smiles, suggesting a natural alignment with the therapeutic process. Patients who are ``Resistant to change'' tend to smile more and more intensely. These smile patterns are mirrored by the therapist, indicating an effort from the therapist to foster interpersonal relations with ``Resistant to change'' patients, likely to enhance the therapeutic outcome.

\paragraph{Alignment}
The alignment between the therapist and the patient tends to grow throughout the conversation \textbf{(RQ1)}. More expressions are reused in the second half than in the first half. This alignment is also more potent when the patient is ``Ready to change'' and ``Receptive'' \textbf{(RQ2)}. The amplitude of the patient's and therapist's bust is more aligned when the patient is ``Ready to change'' and ``Receptive'', and more expressions are reused. Such alignment emerges as the goals of the therapist and the patient are also more aligned. A high verbal alignment can also be a sign of engagement \cite{becker2006concept}. These results can, therefore, be a sign that ``Ready to change'' and ``Receptive'' patients are more engaged in the conversation with the therapist as their initial goal (changing their behavior) is validated by the conversation.

\section{Conclusion}
This paper presents EMMI, composed of two publicly available corpora, AnnoMI and MID, for which we have added multimodal annotations. We analyze these annotations to extract pertinent behavior for developing a virtual agent performing motivational interviews emphasizing social behaviors. We highlight differences in behavior for both the therapist and the patient during the conversations and depending on the change talk conducted by the patient. Our analysis shows that patients can be clustered into three groups of patients (``Ready to change'', ``Resistant to change'', and ``Receptive'') whose behavior differs significantly. The therapist also exhibits different behaviors depending on the type of interlocutor. This shows the importance of virtual interviewers to adapt their behaviors depending on the type of their human patient and their current behavior. In future work, identifying and characterizing these patient types could be a foundation for developing a multimodal dialog to deliver customized MI interventions. The insights gained from understanding how patients evolve in their communication style, behavior expressivity, social and empathic behaviors, and alignment throughout the conversation can inform the design of an intelligent virtual MI interviewer that could dynamically adapt its communication strategies, voice, and content based on the identified type of patient, thus enhancing the effectiveness of MI. 

 \section*{Acknowledgment}

 This work was partially funded by the ANR-DFG-JST Panorama and ANR-JST-CREST TAPAS (19-JSTS-0001-01) projects.

\section*{Ethical Impact Statement}

While the primary aim of our research is not to replace therapists but rather to extend support and accessibility to underserved populations, the personalized adaptability we propose could inadvertently open avenues for manipulation. Introducing an empathetic virtual therapist also raises concerns regarding human-agent attachment, which necessitates careful consideration. In our final application, we are committed to transparently informing participants that they are engaging with a virtual therapist, one that is fallible, may make mistakes, and lacks emotional capacity. It is crucial to emphasize that the agent is not a substitute for human-to-human therapy.

The annotations conducted on videos sourced from YouTube feature actors and do not include any personally identifiable information, thus safeguarding privacy. While the use of actors may introduce a bias, it also allows for privacy protection.

Acknowledging that our data primarily consists of interactions with American participants introduces a potential bias in dialog acts classification results, particularly concerning vocabulary usage. To mitigate this risk, we aim to incorporate examples from other dialects, such as British or African-American Vernacular English, to enhance the classifier's adaptability across diverse linguistic contexts.
The proposed dialog act classifier will be used as a natural language understanding module within human-agent conversations. However, it is imperative to recognize that automatic classifications can yield errors, particularly in the sensitive context of therapy, where mistakes can have significant repercussions. Consequently, outputs from the classifier should be handled with utmost care. This is also true of the proposed separation of patients into types.
In contrast to prevailing methodologies, we have opted to utilize the open-source Mistral model instead of OpenAI's GPT. This enables cost-free usage, open-source availability, and offline deployment. Given the importance of maintaining patient privacy in the context of utterance classification in therapy, the ability to operate the model offline is essential.

% All annotations and code will be made publicly available to facilitate reproducibility.
\bibliographystyle{IEEEtran}
\bibliography{conference_101719}

\end{document}